\documentclass{article}

\usepackage[utf8]{inputenc} 
\usepackage[T1]{fontenc}    
\usepackage{hyperref}       
\usepackage{url}            
\usepackage{booktabs}       
\usepackage{amsfonts}       
\usepackage{nicefrac}       
\usepackage{microtype}      
\usepackage{xcolor}         
\usepackage{subcaption}
\usepackage[vlined]{algorithm2e}
\usepackage{authblk}

\usepackage[disable]{todonotes}
\usepackage{amsmath}

\SetKwFor{RepTimes}{repeat}{times}{end}
\SetKwInOut{Input}{input}

\title{Increasing Performance And Sample Efficiency With Model-agnostic Interactive Feature Attributions}

\author[1]{
  Joran Michiels \thanks{Corresponding author}}
\author[1,2]{
 	Maarten De Vos}
\author[1]{
 	Johan Suykens}
\affil[1]{ESAT-STADIUS, KU Leuven}
\affil[2]{Department of Development and Regeneration, KU Leuven}
\affil[ ]{\textit {\{joran.michiels,maarten.devos,johan.suykens\}@kuleuven.be}}

\begin{document}

\maketitle

\begin{abstract}
  	Model-agnostic feature attributions can provide local insights in complex ML models. If the explanation is correct, a domain expert can validate and trust the model's decision. However, if it contradicts the expert's knowledge, related work only corrects irrelevant features to improve the model. To allow for unlimited interaction, in this paper we provide model-agnostic implementations for two popular explanation methods (Occlusion and Shapley values) to enforce entirely different attributions in the complex model. For a particular set of samples, we use the corrected feature attributions to generate extra local data, which is used to retrain the model to have the right explanation for the samples. Through simulated and real data experiments on a variety of models we show how our proposed approach can significantly improve the model's performance only by augmenting its training dataset based on corrected explanations. Adding our interactive explanations to active learning settings increases the sample efficiency significantly and outperforms existing explanatory interactive strategies. Additionally we explore how a domain expert can provide feature attributions which are sufficiently correct to improve the model.
\end{abstract}

\section{Introduction}

With machine learning models becoming more complex and their application more widespread, pressure to understand the model's decisions is increasing. This has resulted in the proposal of many approaches to increase insights into the model. To not compromise the performance of complex models and allow for straight-forward model comparison, model-agnostic post-hoc explanations are preferred. Their generality and ease of use has made them a very popular choice to validate the decision of the model and/or increase trust. But what if the explanation is actually wrong according to the expert? This is a perfect opportunity to improve the model by changing its explanation, especially when data is scarce or during active learning when labelling is difficult. In this work we show how model-agnostic post-hoc explanations can be made fully interactive, in particular how corrections to these explanations can be enforced in any model by modifying the training data. This extends the existing state-of-the-art \cite{teso2019explanatory} which only allows limited correction of irrelevant features. Besides, this work creates a direct connection between feature attributions and model improvement, providing a use case to test human interpretation of feature importance in further work.

The specific contributions of this paper are:

\begin{itemize}
\item a model optimization setup using interactive explanations, incorporating standard supervised learning and the previously proposed active learning setup \cite{teso2019explanatory, schramowski2020making};
\item implementations of two novel interactive feature attribution methods, respectively for Occlusion \cite{zeiler2014visualizing} and the much used SHAP explanations \cite{lundberg2017unified}, that significantly improve model performance compared to related work \cite{teso2019explanatory} and standard active learning (increasing sample efficiency by more than 50\% in all experiments); 
\item theoretical validation of our methods and comparison with related work \cite{teso2019explanatory};
\item a novel experiment exploring how domain knowledge can be encoded in feature attributions which can be successfully used to improve model performance using our approach.
\end{itemize}
 \todo{provide numbers, label}

This paper will first discuss the necessary background and related work. Next we introduce our interactive feature attributions and provide some theoretical validation. Finally we discuss our experiments and conclude with an outlook on future work. 

\section{Background}
Interactive explanations have mostly appeared in the context of active learning \cite{teso2019explanatory}. The main goal of \textit{active learning} is reducing the amount of labelling a domain expert has to perform. This is achieved by only asking the expert to label the samples which are difficult to the model, e.g. in the context of classification samples who are close to the decision boundary are probably more interesting to the model. We denote the set of interesting samples as the \textit{query}. The procedure is repeated iteratively to achieve maximal sample efficiency. For a complete overview of active learning we refer to \cite{settles2009active}. \cite{teso2019explanatory} proposed the addition of explanations in the active learning loop to increase trust (of the expert) in the model and further increase sample efficiency. An explanation accompanies each interesting sample. Therefore it is \textit{local} in contrast to global, which attempts to explain the whole model. 

In this paper the focus will be on a specific type of local explanations, \textit{feature attributions}. They explain the model output $f(\boldsymbol{x})$ of a single sample $\boldsymbol{x}$ (note the bold font) by assigning an attribution $R_i$ to each feature value $\boldsymbol{x}_i$. The attributions can be interpreted as signed feature importances or local feature effects on the model output. They can be either model-specific or model-agnostic. Model-specific methods may require certain model characteristics (differentiability for Integrated Gradients \cite{sundararajan2017axiomatic}) or can be efficient implementations of a model-agnostic concept (SHAP values for trees \cite{lundberg2020local}). Model-agnostic implementations treat the model as a black-box and generate explanations through sampling it.

A simple model-agnostic attribution is Occlusion \cite{zeiler2014visualizing}: 
\begin{equation}
	R_i = f(\boldsymbol{x}) - f([\boldsymbol{x}_{\bar{i}}, b_i]),
	\label{eq:occlusion}
\end{equation}
with $b$ a background input and $\bar{i}$ all features except $i$. This background \todo{include simple example here} input is an important design choice in many model-agnostic implementations and can be seen as the baseline to which we compare the importance of the feature value. Equation (\ref{eq:occlusion}) can be interpreted as the change in model output if feature value $\boldsymbol{x}_i$ becomes known. One could imagine other attributions that are equivalent:
\begin{equation}
	R_{i, S} = f([\boldsymbol{x}_{S_i}, b_{\bar{S}_i}]) - f([\boldsymbol{x}_S, b_{\bar{S}}]), 
\end{equation}
with $S$ a subset of the features, $S_i$ the same set including $i$ and $\bar{S}$ all features except the ones in $S$. Considering an ordering $p$ in which each feature gets known,  $R_{i, S}$ is the attribution if feature $i$ becomes known after the features in $S$ where the unknown values are replaced by corresponding values from the background $b$. If it is unclear which ordering $p$ (or which corresponding set of features $S^p$) is the best one, the average over all $p \in P$  can be used:
\begin{equation}
	R_i =\frac{1}{|P|}\sum_{p} f\left(\left[\boldsymbol{x}_{S_i^p}, b_{\bar{S}^p_i}\right]\right)-f\left(\left[\boldsymbol{x}_{S^p}, b_{\bar{S}^p}\right]\right).
	\label{eq:shapleyimplementation}
\end{equation}
These are known as SHAP values (Shapley additive explanations) \cite{lundberg2017unified}. They are the only feature attributions satisfying a desirable set of properties including \textit{local accuracy}:
\begin{equation}
	f(\boldsymbol{x}) = f(b) + \sum_{i=1}^M R_i,
\end{equation}
meaning that the feature attributions ($M$ is total number of features) sum to the difference in the model output between the explained sample $\boldsymbol{x}$ and the background sample $b$.

Both Occlusion and SHAP can also be averaged over many background inputs. In fact, the original SHAP value definition computes the expectation of $R_i$ over all $b$ \cite{lundberg2017unified}. Our paper actually uses so-called Baseline Shapley values \cite{sundararajan2020many}. As will become clear later, assuming a single background input allows for a more straight-forward interactive implementation.

Lastly, note that for linear models $f(x) = \sum_i w_i x_i + w_0$ Occlusion and SHAP values give the same attribution namely:
\begin{equation}
	R_i =  w_i(\boldsymbol{x}_i-b_i).
	\label{eq:linear_attribution}
\end{equation}

\section{Related work}
In \cite{teso2019explanatory, schramowski2020making} interactive local post-hoc explanations were explored for the first time. \cite{teso2019explanatory} introduces the framework of \textit{explanatory interactive learning} (XIL) combining active learning with explanations. Our proposed methods do not distinguish between regular supervised learning and active learning, although the active learning seems a natural setting to exploit our methods.

 \cite{teso2019explanatory} proposes a model-agnostic setup named CAIPI. If the explanation of a sample $\boldsymbol{x}$ wrongly identified a subset of features $C$ as relevant (as per expert feedback), the data is augmented with so-called counterexamples of the form $([\boldsymbol{x}_{\bar{c}}, b_{c}], \boldsymbol{y} )$ for every $c\in C$ with $\boldsymbol{y}$ the corrected target and with $b$ either a random sample, fixed alternative sample or another sample from the training dataset. They show how their counterexamples can easily correct models which are fooled by features (confounders) who are wrongly correlated with the output, also known as the `right for the wrong reason' case or the \textit{Clever Hans} scenario \cite{schramowski2020making}. Through classification experiments on data with known irrelevant features they show how their approach significantly improves the explanations of the models w.r.t. to standard interactive learning. Our approach allows for correcting all feature attribution in stead of only correcting irrelevant features. We will compare their method with ours in a series of experiments. While \cite{teso2019explanatory} provides results using LIME explanations \cite{ribeiro2016should}, their approach can work with any type of explanation method. For maximum comparability, we implement their method with the same explanations as our approach.

Other similar work on specific models (thus not comparable to our model-agnostic approach) includes \cite{ross2017right} where input gradients are constrained to improve differentiable models effectively and efficiently. \cite{zaidan2007using, small2011constrained} explore how domain knowledge can be used to constrain a support vector machine in a supervised learning setting while \cite{kulesza2015principles} examines how experts can improve a Naïve Bayes model. For more related work, we refer to \cite{teso2022leveraging}.

\cite{schramowski2020making} further investigates XIL in the Clever Hans scenario and successfully uses the method of \cite{ross2017right} on a plant phenotyping task. In \cite{friedrich2023typology} several methods are combined in a framework and benchmarked on specific tasks to correct the shortcut behaviour from Clever Hans models.

\section{Improving the model through interactive feature attributions}
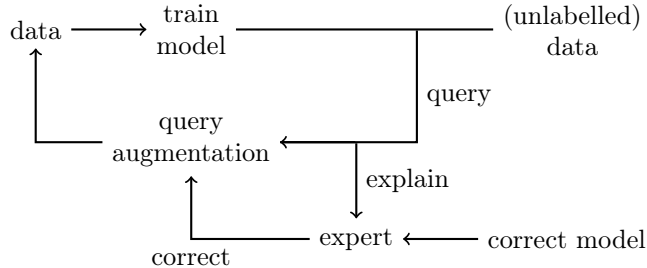
\begin{figure}
\begin{center}
	\begin{tikzpicture}[thick]
		\node (model) at (0,0) [align=center] {train \\ model};
		\node (data) [left=of model] {data};
		\node (a) at (3,0) [inner sep=0,minimum size=0] {};
		\node (all) [right=of a, align=center] {(unlabelled) \\ data};
		\node (aug) at (0,-1.5)[align=center]{query \\ augmentation};
		\node (b) [inner sep=0,minimum size=0,right=of aug] {};
		\node (expert) [below=of b] {expert};
		\node (corr) [right=of expert] {correct model};
		
		\draw[->] (data) -- (model);
		\draw (model) -- (all);
		\draw (a) |- node[pos=0.3, right] {query} (aug);
		\draw[->] (b) -- (aug);
		\draw[->] (aug) -| (data);
		\draw[->] (b) --node[midway, right] {explain} (expert);
		\draw[->] (corr) -- (expert);
		\draw[->] (expert) -| node[below] {correct} (aug);
	\end{tikzpicture}
\end{center}
\caption{Model optimization setup with interactive explanations.}
\label{fig:setup}
\end{figure}

We propose the model optimization setup in Figure \ref{fig:setup} where a model can be iteratively improved through interactive explanations of selected samples from the complete dataset (including unlabelled). The model is first trained on the available labelled training data, whereafter an interesting query is selected and explained to the expert. For each sample of the query, the expert is allowed to correct the predicted label (note that if the sample was part of the labelled dataset, its correct label is already in the training data) and fully change its explanation.  

It is assumed that a data generating process exists and we denote the model
mimicking this data generating process perfectly on the available features as a
\textit{correct model} $f^*$. Note that this model might not achieve perfect accuracy on the test set, even without the presence of noise, because not all necessary features are available to make the model deterministic. On the other hand, multiple correct models might exist if more features are available to the model than necessary. In practice a correct model is unknown (otherwise machine learning would not be necessary) but we assume the domain expert has some knowledge of it which can be expressed through the chosen local explanation method.

To keep our method widely applicable and model-agnostic, the correct explanations are enforced on the model by retraining it on augmented data. 

\subsection{Query augmentation}
For each sample $\boldsymbol{x}$ in the query, new samples are distilled from its (corrected) target $\boldsymbol{y}$ and its correct feature attributions. Knowledge of background input $b$ is also assumed. Unless mentioned otherwise, the augmentation is identical for regression ($f(\cdot), \boldsymbol{y} \in \mathbb{R}$) and classification ($f(\cdot), \boldsymbol{y} \in \{0,1\}$). 

Correct explanations $R^*_i$ for a sample $\boldsymbol{x}$ can be enforced in model $f$ by retraining $f$ so that its explanations $R_i$ are equal to $R^*_i$. In the case of Occlusion, by Equation (\ref{eq:occlusion}) we know that:
\begin{equation}
	\begin{split}
		R_i &= R^*_i  \\
		f(\boldsymbol{x}) - f([\boldsymbol{x}_{\bar{i}}, b_i]) &= R^*_i.
	\end{split}
	\label{eq:output_change}
\end{equation} 
Combining this constraint with $f(\boldsymbol{x})$ = $\boldsymbol{y}$ (either the target was already known or has been corrected by the expert) we get:
\begin{equation}
	f([\boldsymbol{x}_{\bar{i}}, b_i]) = \boldsymbol{y} - R^*_i 
\end{equation}
Thus the sample $([\boldsymbol{x}_{\bar{i}}, b_i], \boldsymbol{y}-R^*_i)$ should be added to the training dataset. Note the similarity with the state-of-the-art \cite{teso2019explanatory}, if $R^*_i=0$ our sample augmentations are identical to theirs.

Repeating the procedure above for SHAP values (this time for all $i$), we get the following constraints:
\begin{equation}
	\left\{
	\begin{array}{l} \forall i:
		\frac{1}{|P|}\sum_{p} f\left(\left[\boldsymbol{x}_{S_i^p}, b_{\bar{S}^p_i}\right]\right)-f\left(\left[\boldsymbol{x}_{S^p}, b_{\bar{S}^p}\right]\right) = R^*_i  \\
		f(\boldsymbol{x})$ = $\boldsymbol{y} .
	\end{array}
	\right.
	\label{eq:shapley_constraints}
\end{equation}
This set of equations is underdetermined so extra constraints have to be imposed. A sensible choice would be enforcing the same attribution for every subset. In the Appendix we show how this leads to training samples of the form  $([\boldsymbol{x}_{\bar{S}_i}, b_{S_i}], \boldsymbol{y}-\sum_{j \in S_i}{R^*_j})$ or $([\boldsymbol{x}_{\bar{S}_i}, b_{S_i}], H(\boldsymbol{y}-\sum_{j \in S_i}{R^*_j} -0.5))$, with $H$ the Heaviside function, in the case of classification.

Note that in classification settings it is customary to compute explanations on the decision function or class probability to get continuous and more fine-grained feature importances \cite{lundberg2018explainable}. For example, for logistic regression it is easy to see why this is preferred: if we compute the explanation on the log odds, we are essentially explaining a linear model which has very intuitive (see Equation (\ref{eq:linear_attribution})) attributions. However in the context of model-agnostic interactive explanations it makes little sense, since most classifiers require samples with discrete targets and domain experts tend to provide a corrected class and not a class probability.

\subsection{Theoretical validation}

In the case of linear models, we can examine how the described query augmentation allows to correct the model. Consider a correct model $f^*$ and incorrect model $f$,
\begin{equation}
	f^*(x) = \sum_i w^*_i x_i + w^*_0 \quad f(x) = \sum_i w_i x_i + w_0
\end{equation}
We aim to correct $f$ by correcting a number of local attributions with background input $b$. For query $\boldsymbol{x}$, correct target $\boldsymbol{y}$ and feature attribution $R^*_k$ the sample $([\boldsymbol{x}_{\bar{k}}, b_k], \boldsymbol{y}-R^*_k))$ is added to the training data. Including the original (corrected) sample, model $f$ should adhere to the following constraints:
\begin{equation}
	\left\{
	\begin{array}{l}
		\boldsymbol{y} = \sum_{i} w_i \boldsymbol{x}_i + w_0 \\
		\boldsymbol{y}- R^*_k = \sum_{i \neq k} w_i \boldsymbol{x}_i + w_k b_k + w_0 .
	\end{array}
	\right.
\end{equation}
Subtracting the constraints, we find that $R^*_k =  w_k(\boldsymbol{x}_k-b_k)$, so if we aim to correct $w' \rightarrow w^*_k$, $R^*_k \equiv w^*_k(\boldsymbol{x}_k-b_k)$. This is the SHAP and Occlusion attribution for linear models. \todo{maybe improve the wording here} This means that a single augmentation can correct the corresponding model parameter. One correct explanation can completely correct a linear model. As explained above, the approach from \cite{teso2019explanatory} correspond to setting $R^*_k=0$ and is thus only correct if $w^*_k=0$, otherwise it incorrectly pushes $w_k \rightarrow 0$. 

For non-linear models, the approach with Occlusion comes down to adding an extra correct sample to the training data, and is thus expected to be beneficial. \todo{maybe discuss out of domain samples in limitation} In case of SHAP, since the extra samples were produced after imposing constraints, they do not necessarily correspond to correct samples from the correct model. In the Clever Hans scenario  \cite{schramowski2020making} a confounded feature $c$ has an attribution $R_c \neq 0$. Since $\boldsymbol{x}_c$ will not be meaningful, an expert should set $R_c^*=0$, making our method equivalent to \cite{teso2019explanatory}.
\subsection{Implementation}

To eliminate any bias towards the newest queries, the model is retrained on all data (including the augmented data) every iteration. To maintain a balance between adding new samples and adding corrections, for each correction, we propose to add the new sample to the dataset (thus oversampling it). In the case of Occlusion we also allow for incomplete explanations, if the expert has partial domain knowledge. Complete implementations are in Algorithm \ref{correct_query_occ} and \ref{correct_query_shap}. Note that if we set $R_i = 0$ for all $i$ who are irrelevant, Algorithm \ref{correct_query_occ} can and will be used as an implementation for CAIPI \cite{teso2019explanatory}.

\begin{figure}
\begin{minipage}{0.46\textwidth}
	
	\RestyleAlgo{ruled}
	\begin{algorithm}[H]
		\Input{$X_{tr}, Y_{tr}, X_{query}, b$}
		\For{$\boldsymbol{x} \in X_{query}$}{
			\If{$\boldsymbol{x}$ is unlabelled}
			{get $\boldsymbol{y}$ from expert\;}
			get $R^*$ from expert\;
			\For{$R^*_i \in R^*$}{
				\If{$R^*_i$ is available}{
					$\boldsymbol{x}_{aug} \leftarrow \boldsymbol{x}$\;
					$\boldsymbol{x}_{aug,i} \leftarrow b_i$\;
					$\boldsymbol{y}_{aug} \leftarrow \boldsymbol{y} - R^*_i$ \;
					add $\boldsymbol{x}_{aug}, \boldsymbol{y}_{aug}$ to $(X_{tr},Y_{tr})$\;
					add $\boldsymbol{x}, \boldsymbol{y}$ to $(X_{tr},Y_{tr})$\;
			}}	
		}
		
		retrain model on $(X_{tr}, Y_{tr})$\;
		\caption{Augmentation Occlusion}%
		\label{correct_query_occ}%
	\end{algorithm}%
\end{minipage}
\hfill
\begin{minipage}{0.5\textwidth}
	
	\RestyleAlgo{ruled}
	\begin{algorithm}[H]
		\RestyleAlgo{ruled}
		\Input{$X_{tr}, Y_{tr}, X_{query}, b, K$}
		\For{$\boldsymbol{x} \in X_{query}$}{
			\If{$\boldsymbol{x}$ is unlabelled}
			{get $\boldsymbol{y}$ from expert\;}
			get $R^*$ from expert\;
			\RepTimes{K}{
				$p \leftarrow Permutation(\{1,2, \dots M\})$\;
				$\boldsymbol{x}_{aug} \leftarrow \boldsymbol{x}$\;
				$\boldsymbol{x}_{aug,S^p_i} \leftarrow b_{S^p_i}$\;
				$\boldsymbol{y}_{aug} \leftarrow \boldsymbol{y} - \sum_{j \in S^p_i} R^*_j$ \; 
				append $\boldsymbol{x}_{aug}, \boldsymbol{y}_{aug}$ to $(X_{tr},Y_{tr})$\;
				append $\boldsymbol{x}, \boldsymbol{y}$ to $(X_{tr},Y_{tr})$\; 
			}	
		}
		
		retrain model on $(X_{tr}, Y_{tr})$\;
		\caption{Augmentation SHAP}%
	    \label{correct_query_shap}%
	\end{algorithm}%
\end{minipage}
\end{figure}

\section{Experiments}
The goals of these experiments are to confirm our theoretical validation; to show the generality of our methods in different training tasks (linear and non-linear complex models, classification and regression); and examine the advantages over related work \cite{teso2019explanatory} on interactive explanations.

We compare our proposed methods which allow for full correction of the explanation (denoted by \textit{Interactive Occlusion} or \textit{Interactive SHAP}) against the state-of-the-art which only correct irrelevant features (denoted by \textit{CAIPI}) and against regular active learning (only correcting the targets, denoted by \textit{Baseline}). In the experiments, the interesting queries are selected by random. As a background value $b$ either the average or the median (in case of discrete features) of the complete dataset is used. To fairly compare all strategies, the number of extra (augmented) samples per iteration is the same for all strategies.  Our Occlusion implementation adds $2|R^*|$ samples per sample $\boldsymbol{x}$ per iteration (see Algorithm \ref{correct_query_occ}), thus in the results of standard active learning, $\boldsymbol{x}$ is oversampled $2|R^*|$ times. Similarly, in the results with SHAP values, $K=|R^*|$ in Algorithm \ref{correct_query_shap}. If $R^*$ is not completely determined (some attributions are missing, this is the case for CAIPI), we again oversample $\boldsymbol{x}$ to make sure the amount of augmented samples is the same. 

As is customary in active learning, we simulate the domain expert by querying a completely labelled dataset. Correct explanation are also simulated in related work \cite{teso2019explanatory, ross2017right}. Our first set of experiments continues this practice, testing our first hypotheses that \textbf{correct feature attributions can be used to improve models}. In a final experiment the authors' (basic) domain knowledge is encoded into approximately correct explanations, shedding light on our second hypotheses: \textbf{domain experts can supply feature attributions that are sufficiently correct}.

All experiments were run on a server with 24 CPU nodes within an hour. Code for our implementations and all experiments is available on request. We enjoyed using the Python packages \texttt{scikit-learn} \cite{scikit-learn} and \texttt{shap} \cite{lundberg2017unified} (solely KernelSHAP is used).

\subsection{Correct feature attributions can be used to improve models}
\begin{figure}
	\begin{subfigure}{0.48\textwidth}
		\centering
		\includegraphics[width=\textwidth]{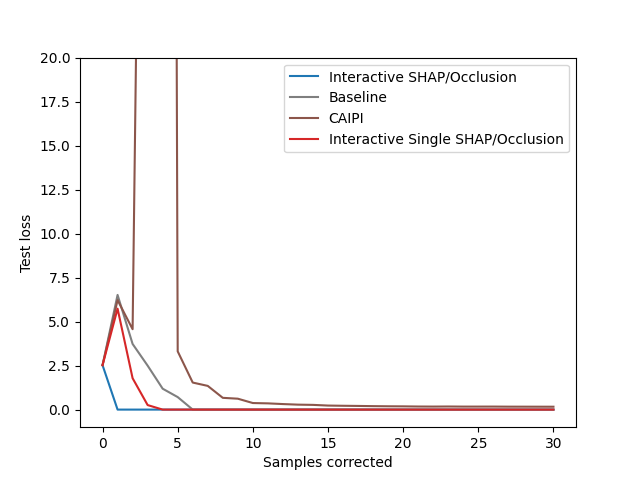}
		\caption{}
	\end{subfigure}
	\begin{subfigure}{0.48\textwidth}
		\centering
		\includegraphics[width=\textwidth]{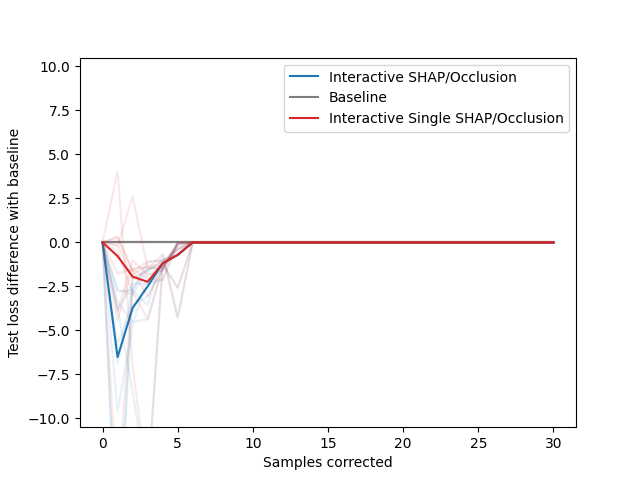}
		\caption{}
	\end{subfigure}
	\caption{Comparison of test loss evolution for different active learning strategies. The experiment was repeated for five random linear models and five randomly shuffled datasets per model. (a) shows the average test loss for our strategies (\textit{Interactive SHAP/Occlusion} and \textit{Interactive Single SHAP/Occlusion}), the current state-of-the-art \cite{teso2019explanatory} (\textit{CAIPI}) and the standard approach without explanations (\textit{Baseline}). To determine the significance Figure (b) shows the average of the per run difference in test loss of our approaches w.r.t. the baseline. The faded lines are the actual differences.}
	\label{fig:linear}
\end{figure}

These experiments take the expert out of the equation using a known data generating model. First, to confirm our theoretical validation, a known linear model with five uniformly generated coefficients $w^*_i \in [-1, 1]$ is considered which is used to generate noiseless targets for normal distributed Gaussian data:
\begin{equation}
	\begin{gathered}
		x_i \sim \mathcal{N}(0,1) \ \text{for} \ i = 1\dots5 \\
		y = \sum_i w^*_i x_i + w^*_0.
	\end{gathered}
	\label{eq:linear_generation}
\end{equation}
An untrained linear model with all coefficients zero is then iteratively corrected by considering one queried sample per iteration. The evolution of the test loss on 100 hold-out samples for different strategies is shown in Figure \ref{fig:linear}. As already mentioned in the background SHAP and Occlusion attributions are equivalent for linear models. Algorithm \ref{correct_query_occ} is used to augment every sample. To provide a comparison with the state-of-the art \cite{teso2019explanatory}, the least important feature $u$ of the queried sample is considered irrelevant, i.e. only $R^*_u = 0$ in Algorithm \ref{correct_query_occ}. This can be directly compared to only setting $R^*_u$ to its correct SHAP/Occlusion attribution (denoted by \textit{Interactive Single SHAP/Occlusion}).

\begin{figure}
	\begin{subfigure}{0.48\textwidth}
		\centering
		\includegraphics[width=\textwidth]{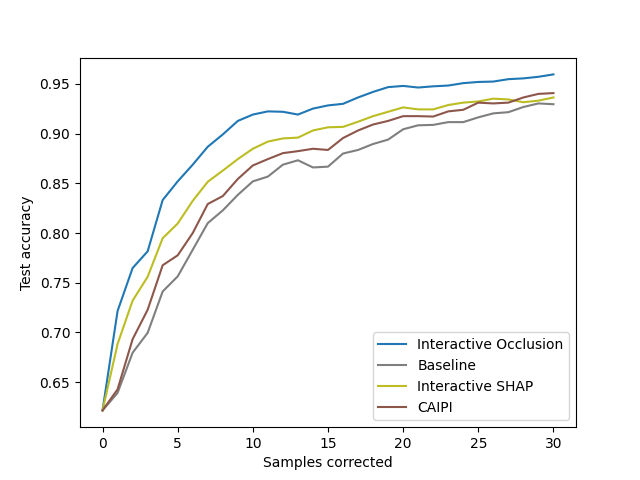}
		\caption{}
	\end{subfigure}
	\begin{subfigure}{0.48\textwidth}
		\centering
		\includegraphics[width=\textwidth]{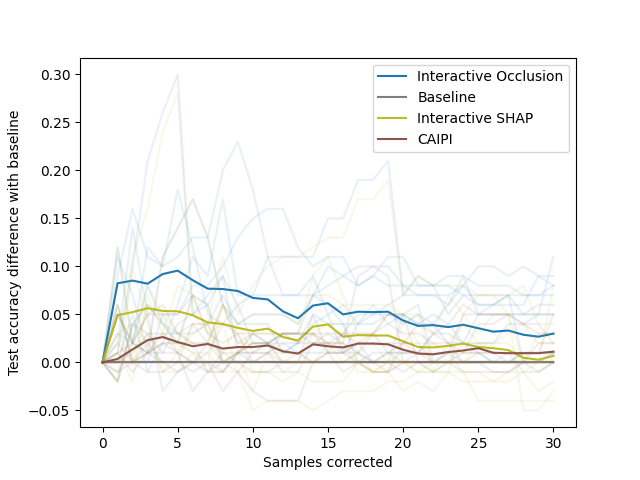}
		\caption{}
	\end{subfigure}
	\caption{Comparison of test accuracy evolution for different active learning strategies. The experiment was repeated for five random logistic regression models and five randomly shuffled datasets per model. (a) shows the average test accuracy for our strategies (\textit{Interactive Occlusion} and \textit{Interactive SHAP}), the state-of-the-art \textit{CAIPI} and the standard approach without explanations (\textit{Baseline}) . To determine the significance, Figure (b) shows the average of the per run difference in test accuracy of all approaches w.r.t. the baseline. The faded lines are the actual differences.}
	\label{fig:logistic}
\end{figure}

A similar experiment is repeated for a logistic regression model. Again, samples are generated according to Equation (\ref{eq:linear_generation}) but this time $y = H(\sum_i w^*_i x_i + w^*_0)$ with $H$ the Heaviside function. In this case, initially the model is trained on two random samples of different classes (since the training dataset has to have two different classes) whereafter each iteration a new sample is added and corrected. In this setting CAIPI, which considers all features with Occlusion attribution $R_k = 0$ to be irrelevant, should be directly compared to our methods. The evolution of the test accuracy on 100 hold-out samples is shown in Figure \ref{fig:logistic}.

\begin{figure}
	\begin{subfigure}{0.48\textwidth}
		\centering
		\includegraphics[width=\textwidth]{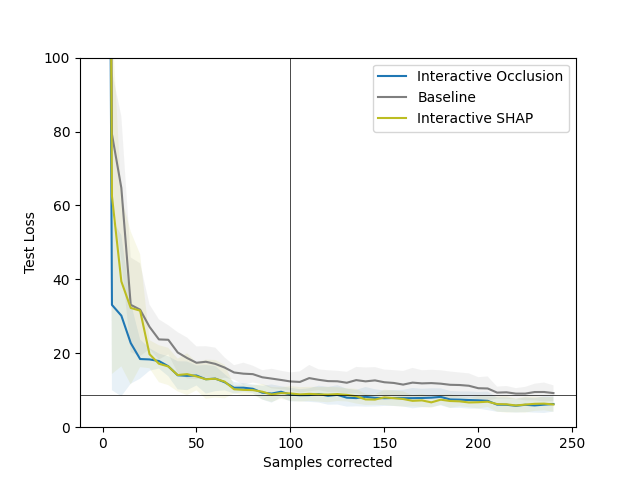}
		\caption{}
	\end{subfigure}
	\begin{subfigure}{0.48\textwidth}
		\centering
		\includegraphics[width=\textwidth]{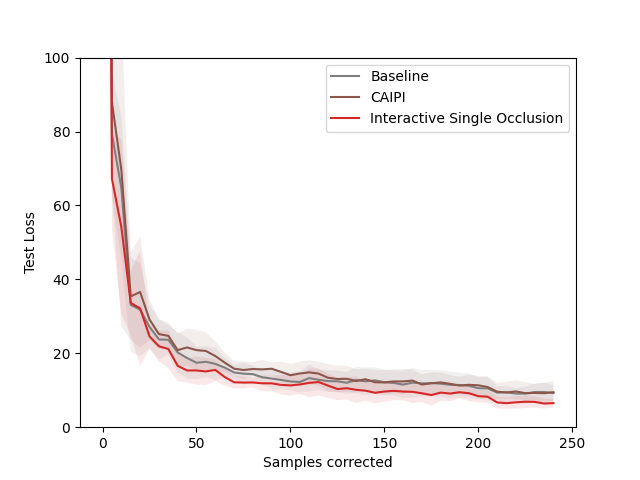}
		\caption{}
	\end{subfigure}
	\caption{Comparison of test loss evolution for different active learning strategies on the Boston Housing dataset. The experiment was repeated for five randomly shuffled datasets. (a) shows the average test loss for our strategies (\textit{Interactive Occlusion} and \textit{Interactive SHAP}) and the standard approach without explanations (\textit{Baseline}). (b) shows the average test loss for the current state-of-the-art \cite{teso2019explanatory} \textit{CAIPI} and \textit{Interactive Single SHAP/Occlusion}.}
	\label{fig:boston}
\end{figure}

To show the generality of our method, we next consider a more complex dataset, the Boston Housing dataset \cite{harrison1978hedonic}. In this case, the real data generating model is unknown. To accurately provide correct attributions we train a boosted forest with 10 trees on the complete dataset and use the predicted house prices as the labels, and the correct explanations to iteratively improve an untrained boosted forest (with the same hyperparameters). The evolution of test loss on 253 hold out samples (half of the dataset) is shown in Figure \ref{fig:boston}. This time we consider a random query of 5 new samples each iteration. As in the experiment with linear models, we also include a \textit{Interactive Single SHAP/Occlusion} as an extra comparison with the state-of-the-art.

The Appendix holds additional experiments with different models.
\subsection{Domain experts can supply feature attributions that are sufficiently correct}\label{expert_results}

\begin{figure}
	\begin{subfigure}{0.40\textwidth}
	\begin{algorithm}[H]
		
		\Input{$\boldsymbol{x}, \boldsymbol{y}$}
		\If{$\boldsymbol{y} = 1$}
		{\uIf{$\boldsymbol{x}_{sex} = female$}
			{\If{Occlusion}
			{$R^*_{sex} \leftarrow 1$;}\
			\If{CAIPI}
			{$R^*_{/sex} \leftarrow 0$;}}\
		\uElseIf{$\boldsymbol{x}_{age} < 12$}
			{\If{Occlusion}
			{$R^*_{age} \leftarrow 1$;}\
			\If{CAIPI}
			{$R^*_{/age} \leftarrow 0$;}\
			}}%
	\end{algorithm}%
    \caption{}
	\label{titanic_expert}
	\end{subfigure}
\begin{subfigure}{0.6\textwidth}
	\centering
	\includegraphics[width=\textwidth]{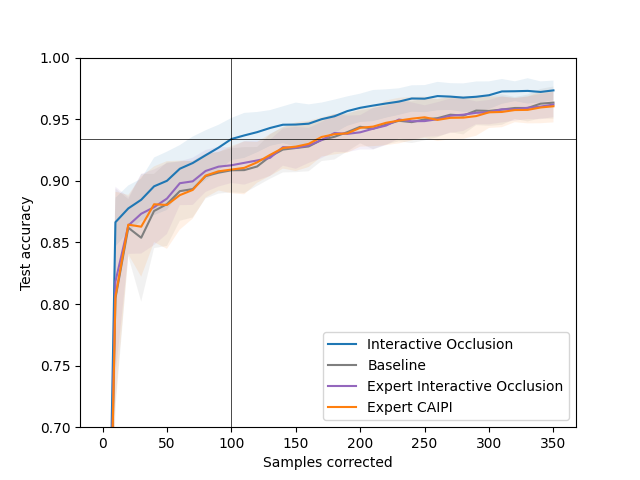}
	\caption{}
	\label{fig:titanic}
\end{subfigure}
	\caption{Figure (a) holds a procedure to encode Titanic domain knowledge in feature attributions. Figure (b) holds the comparison of test accuracy evolution for different active learning strategies on the Titanic dataset. The experiment was repeated for twenty randomly shuffled datasets. The figure shows the average test accuracy for entirely correct explanations (\textit{Interactive Occlusion}), explanations generated by experts (\textit{Expert Interactive Occlusion, Expert CAIPI}) and standard active learning.}
	
\end{figure}

To motivate the practical use of our proposed methods, we show how expert knowledge about the data can lead to an increased performance through interactive feature attributions. The task is to predict the survival of Titanic passengers based on class, sex, age and number of siblings (\textit{SibSp}). To determine an upper bound, the dataset is again simulated as in the above experiment and explanations from the correct model (a boosted forest with 10 trees) are used to improve an untrained boosted forest (same hyperparameters as correct model, trained on two random samples of different classes). While the authors are no experts on the survival of passengers aboard the Titanic, they are familiar with the phrase `women and children first'. This piece of knowledge can be encoded in Occlusion feature attributions: with regard to the known background sample $(class=2,sex=male,age=28,SibSp=0)$, a procedure to set the correct attributions is shown in Figure \ref{titanic_expert}. For the state-of-the-art approach CAIPI \cite{teso2019explanatory} the irrelevant features get an attribution of zero. Figure \ref{fig:titanic} shows the results when correcting a random query of 10 samples each iteration.

\section{Discussion}
\subsection{Correct feature attributions can be used to improve models}
The results with linear models (Figure \ref{fig:linear}) show that our methods significantly outperform the state-of-the-art \cite{teso2019explanatory} and standard active learning on sample efficiency. Also, correcting one complete explanation is indeed sufficient to fully correct a linear model (\textit{Interactive SHAP/Occlusion} goes to zero after one sample). CAIPI diverges outside the figure boundary, most likely because every iteration, theoretically the coefficient associated with the irrelevant feature $w_u$ gets pushed to zero. This is in stark contrast to setting $R_u$ to its correct SHAP/Occlusion attribution: \textit{Interactive Single SHAP/Occlusion} still converges rather rapidly to zero loss. 

With logistic regression we see very similar results, our methods outperform the state-of-the art significantly. Algorithmically speaking, the only difference between \textit{Interactive Occlusion} and \textit{CAIPI} is that our approach also enforces nonzero feature attributions (in the classification setting either -1 or 1). At sample level this results in extra samples containing feature values that change the class, which can play a big role in training classification models efficiently. This also motivates exploring interactive counterfactual explanations (\cite{verma2020counterfactual} is a review). Making the same mistake as CAIPI, \cite{ross2017right} also only enforces zero input gradients.

On the Boston Housing data, our methods also do better (increasing sample efficiency by more than 100\%) than CAIPI, which does worse than the baseline. However, the method does not diverge such as was the case for linear models. This is probably because of the increased complexity of the boosted tree model, meaning extra local data has less effect on the global model. A linear model is fully defined by its explanation (see our theoretical validation). Moreover the approach with SHAP values, while still being significantly better than standard active learning, is worse than using Occlusion attributions (we see the same with logistic regression). While it is argued that the generality and intuitive properties (such as local accuracy) of the SHAP value increase its understandability \cite{lundberg2017unified}, our proposed interactive implementation does not outperform the much more simple Occlusion attribution. This was partly expected because the added complexity of SHAP results in extra constraints to generate the local augmented data, which means the extra samples are not necessarily samples from the correct model. Therefore in the next experiment we will only use Occlusion attributions.

Note that all of the active learning results can be easily interpreted as regular supervised learning results if they are read in the vertical direction for different fixed number of corrected samples. Doing so shows that adding augmented data (according to our implementations) to particular fixed sets of samples always increased the performance of the model.

\subsection{Domain experts can supply feature attributions that are sufficiently correct}
The experiment on the Titanic dataset shows that general knowledge can indeed be encoded in Occlusion feature attributions leading to an increase in performance. Our approach outperforms the state-of-the-art \cite{teso2019explanatory}. It seems more intuitive to provide correct attributions based on the domain knowledge of the Titanic dataset, than to denote the irrelevant features. Enforcing that features are completely irrelevant is too drastic: e.g. although sex might be the most relevant feature, other features such as age can also be relevant, but to a lesser degree.

\section{Limitations}
Most of our experiments use exactly correct attributions (also in \cite{teso2019explanatory}). In reality, domain experts will make mistakes in correcting attributions, still our last experiment in  Subsection \ref{expert_results} shows that even rudimentary domain knowledge is beneficial. While our approach significantly increases sample efficiency, it is unclear if is time efficient: correcting attributions might take more time than simply correcting extra samples. However, usually the expert has a reason to correct a target, if this reason can be efficiently encoded in attributions, it should only add a little extra time. 

\section{Conclusion and future work}

This paper introduced the first interactive feature attributions which allow for complete correction. We theoretically discuss our approach and provide connections with related work. Implementations for two popular attributions (Occlusion and SHAP) significantly improve the sample efficiency and performance of several models in regression and classification tasks compared to standard active/supervised learning and the state-of-the-art. Additionally we provide an experiment that showcases how actual domain knowledge can increase the performance of a trained model through our interactive attributions. In further work we plan to involve a real domain expert to further motivate our approach. Overall, our contribution is widely applicable and could significantly improve model performance when data is scarce and increase the sample efficiency of active learning tasks.

As a rare testable practical use case for feature attributions, this work also opens up the discussion on comparing different feature attributions and their human interpretation. Existing work mostly focuses on theory/intuition (\cite{covert2021explaining} is a recent interesting effort), benchmarks (\cite{tomsett2020sanity} provides an overview of different metrics in the image domain). In reality, explanations are to be used by experts and their evaluation should take this into account, e.g. \cite{jesus2021can} examines how different explanation can influence decision accuracy and time. Another evaluation could be through a task involving interactive explanations. Our results already hint that the current state-of-the-art in local explanations, SHAP values, might perform worse than basic Occlusion attributions. The added complexity of advanced explanation methods might not be preferred in interactive settings. Of course, different interactive implementations for SHAP values exist. Further work could consist on joint optimization of an explanation loss (e.g. distance to the correct explanation) and the regular classification or regression loss, similar to \cite{ross2017right}. Additionally, it would be very interesting to further explore how domain experts can encode their knowledge into different types of explanation. In that case, the intuitive properties of SHAP values such as local accuracy might be useful.

\section*{Acknowledgements}
This project has received funding from the Flemish Government (AI
Research Program) and from the FWO (`Artificial Intelligence (AI) for
data-driven personalised medicine', G0C9623N and `Deep, personalized epileptic seizure detection’, G0D8321N) and Leuven.AI Institute.

\bibliographystyle{unsrt}
\bibliography{refs_neurips}


\appendix

\section{Derivation SHAP implementation}

As mentioned above the correct SHAP values $R^*_i$ for sample $\boldsymbol{x}$ and corrected target $\boldsymbol{y}$ impose the following constraints on model $f$:
\begin{equation}
	\left\{
	\begin{array}{l} \forall i \in F:
		\frac{1}{|P|}\sum_{p} f\left(\left[\boldsymbol{x}_{S_i^p}, b_{\bar{S}^p_i}\right]\right)-f\left(\left[\boldsymbol{x}_{S^p}, b_{\bar{S}^p}\right]\right) = R^*_i  \\
		f(\boldsymbol{x})$ = $\boldsymbol{y} ,
	\end{array}
	\right.
	\label{eq:shapley_constraints}
\end{equation}
with $F=\{1\dots M\}$ the set of all features. From this set of constraints we like to distil extra local data (i.e. extra function values) that can be added to the training data to enforce the correct SHAP values on the model.
However, this set of equations is underdetermined: there are $|S|$ unknowns, $f\left(\left[\boldsymbol{x}_{S}, b_{\bar{S}}\right]\right) \ \forall S\subseteq F$,  but only $M + 1$ constraints. Note that $|S| = 2^M > M+1$ for $M \geq 2$. A sensible choice would be enforcing the same attribution for every subset. This results in the following constraints:
\begin{equation}
	\left\{
	\begin{array}{l}
		\forall i \in F, S \subset F : f\left(\left[\boldsymbol{x}_{S_i}, b_{\bar{S}_i}\right]\right)-f\left(\left[\boldsymbol{x}_{S}, b_{\bar{S}}\right]\right) = R^*_i  \\
		f(\boldsymbol{x})$ = $\boldsymbol{y} .
	\end{array}
	\right.
	\label{eq:solvable_shapley_constraints}
\end{equation}
This set of equations is solvable. To determine $f\left(\left[\boldsymbol{x}_{S}, b_{\bar{S}}\right]\right)$ we can use the first constraint of Equation (\ref{eq:solvable_shapley_constraints}) to arrive at:
\begin{equation}
	f\left(\left[\boldsymbol{x}_{S}, b_{\bar{S}}\right]\right) = f(\boldsymbol{x}) - \sum_{j \in \bar{S}}{R^*_j}. 
\end{equation}
Enforcing the second constraint of Equation (\ref{eq:solvable_shapley_constraints}), we get:
\begin{equation}
	f\left(\left[\boldsymbol{x}_{S}, b_{\bar{S}}\right]\right) = \boldsymbol{y} - \sum_{j \in \bar{S}}{R^*_j}. 
\end{equation}
Thus enforcing correct SHAP values $R^*_i$ results in samples of the form $([\boldsymbol{x}_{S}, b_{\bar{S}}], \boldsymbol{y}-\sum_{j \in \bar{S}}{R^*_j})$. Since there are $2^M$ possible subsets,  we will not use \textit{all} training samples of this form. Instead, equivalent to approximating SHAP values \cite{lundberg2017unified}, a number of feature orderings $p$ will be sampled and the corresponding samples $([\boldsymbol{x}_{S^p_i}, b_{\bar{S^p_i}}], \boldsymbol{y}-\sum_{j \in \bar{S^p_i}}{R^*_j}) \ \forall i$ will be added to the training dataset.

\section{Additional experiments}
To further motivate the generality of our methods, we conduct extra experiments on two other model types.

\subsection{Experiment support vector machines}
\begin{figure}
	\begin{subfigure}{0.48\textwidth}
		\centering
		\includegraphics[width=\textwidth]{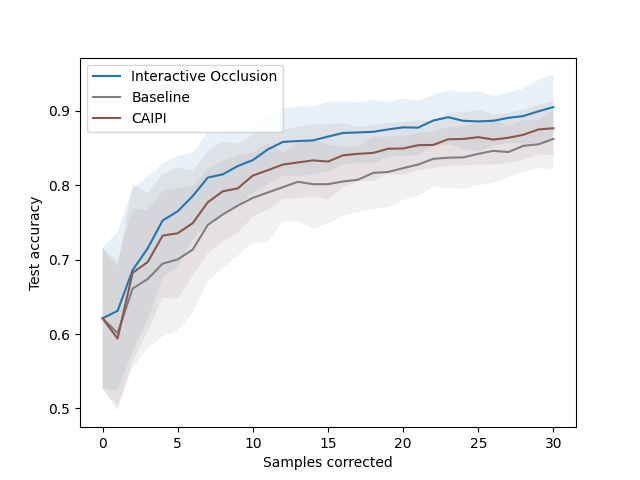}
		\caption{}
	\end{subfigure}
	\begin{subfigure}{0.48\textwidth}
		\centering
		\includegraphics[width=\textwidth]{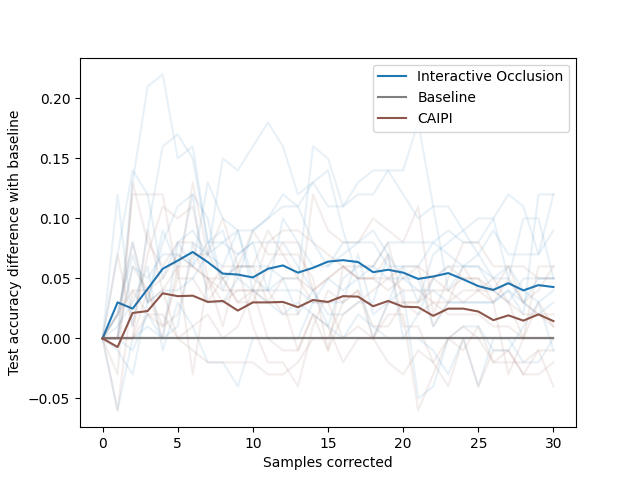}
		\caption{}
	\end{subfigure}
	\caption{Comparison of test accuracy evolution of a support vector machine for different active learning strategies. The experiment was repeated for five random data-generating logistic regression models and five randomly shuffled datasets per model. (a) shows the average test accuracy for our strategy (\textit{Interactive Occlusion}), the state-of-the-art \textit{CAIPI} and the standard approach without explanations (\textit{Baseline}) . To determine the significance, Figure (b) shows the average of the per run difference in test accuracy of all approaches w.r.t. the baseline. The faded lines are the actual differences.}
	\label{fig:svm}
\end{figure}

We use a support vector machine with RBF kernel on the data generated by random logistic regression models (the experiment shown in Figure \ref{fig:logistic}) and achieve the results in Figure \ref{fig:svm}. Our approach outperforms the state-the-art \cite{teso2019explanatory} and the baseline.

\subsection{Experiment multilayer perceptron}
\begin{figure}
	\begin{subfigure}{0.48\textwidth}
		\centering
		\includegraphics[width=\textwidth]{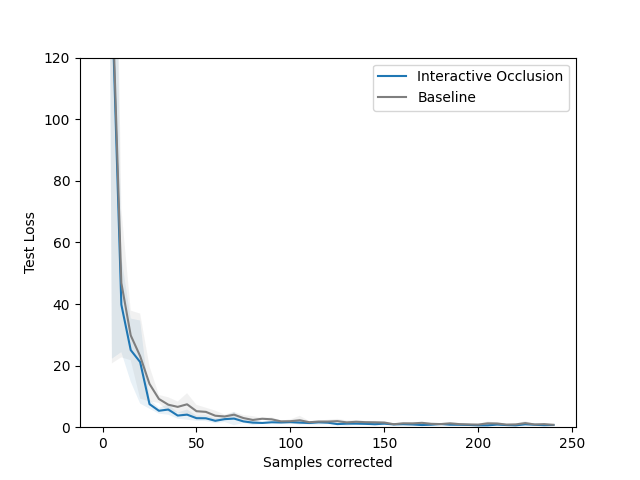}
		\caption{}
	\end{subfigure}
	\begin{subfigure}{0.48\textwidth}
		\centering
		\includegraphics[width=\textwidth]{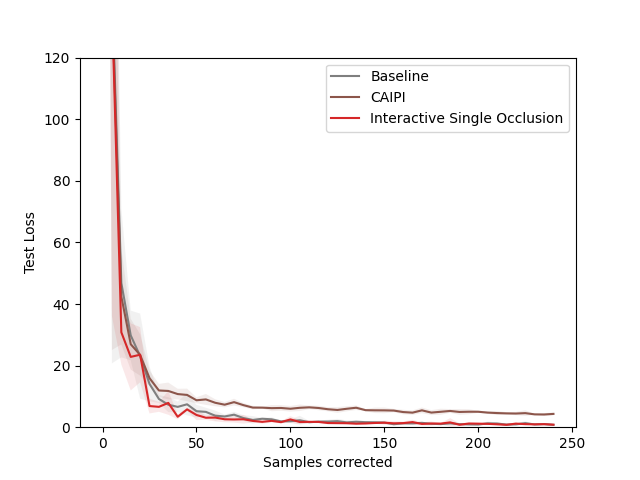}
		\caption{}
	\end{subfigure}
	\caption{Comparison of test loss evolution for different active learning strategies on the Boston Housing dataset. The experiment was repeated for five randomly shuffled datasets. The data-generating model and training model are both multilayer perceptrons. They are trained for 100 epochs. (a) shows the average test loss for our strategy (\textit{Interactive Occlusion}) and the standard approach without explanations (\textit{Baseline}). (b) shows the average test loss for the current state-of-the-art \cite{teso2019explanatory} \textit{CAIPI} and \textit{Interactive Single Occlusion}.}
	\label{fig:boston_100}
\end{figure}

\begin{figure}
	\begin{subfigure}{0.48\textwidth}
		\centering
		\includegraphics[width=\textwidth]{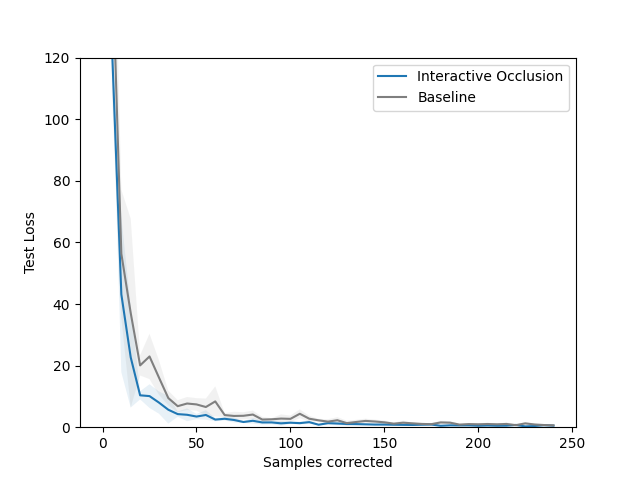}
		\caption{}
	\end{subfigure}
	\begin{subfigure}{0.48\textwidth}
		\centering
		\includegraphics[width=\textwidth]{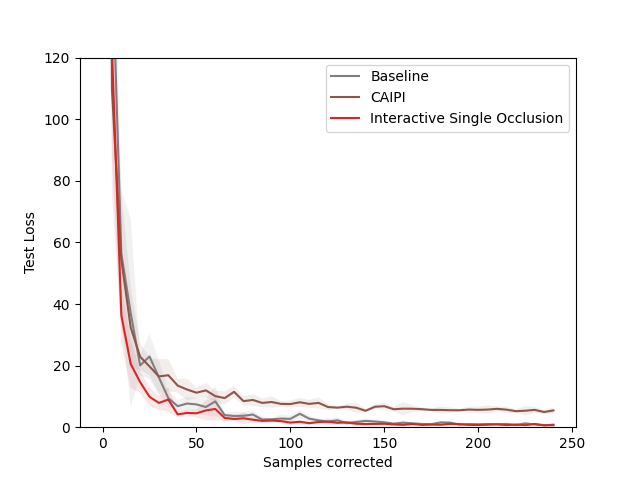}
		\caption{}
	\end{subfigure}
	\caption{Comparison of test loss evolution for different active learning strategies on the Boston Housing dataset. The experiment was repeated for five randomly shuffled datasets. The data-generating model and training model are both multilayer perceptrons. They are trained for 200 epochs. (a) shows the average test loss for our strategy (\textit{Interactive Occlusion}) and the standard approach without explanations (\textit{Baseline}). (b) shows the average test loss for the current state-of-the-art \cite{teso2019explanatory} \textit{CAIPI} and \textit{Interactive Single Occlusion}.}
	\label{fig:boston_200}
\end{figure}

\begin{figure}
	\begin{subfigure}{0.48\textwidth}
		\centering
		\includegraphics[width=\textwidth]{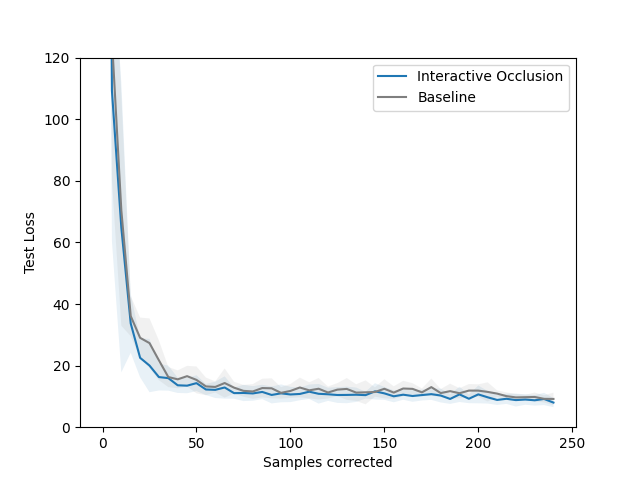}
		\caption{}
	\end{subfigure}
	\begin{subfigure}{0.48\textwidth}
		\centering
		\includegraphics[width=\textwidth]{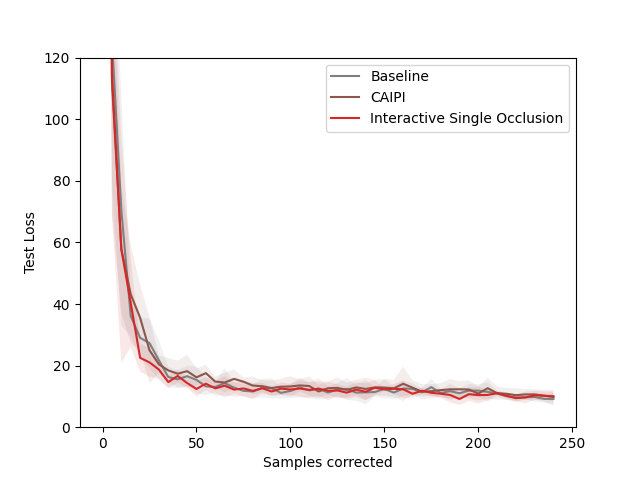}
		\caption{}
	\end{subfigure}
	\caption{Comparison of test loss evolution for different active learning strategies on the Boston Housing dataset. The experiment was repeated for five randomly shuffled datasets. The data-generating model is a boosted forest with 10 trees. The training model is a multilayer perceptron trained for 100 epochs. (a) shows the average test loss for our strategy (\textit{Interactive Occlusion}) and the standard approach without explanations (\textit{Baseline}). (b) shows the average test loss for the current state-of-the-art \cite{teso2019explanatory} \textit{CAIPI} and \textit{Interactive Single Occlusion}.}
	\label{fig:boston_forest}
\end{figure}

Lastly we train a multilayer perceptron with 9 hidden nodes (using stochastic gradient descent) iteratively on the Boston housing dataset \cite{harrison1978hedonic} (similar experiment to Figure \ref{fig:boston}). In Figure \ref{fig:boston_100} and Figure \ref{fig:boston_200} the data generator is a multilayer perception with the same hyperparameters. In Figure \ref{fig:boston_forest}, it is the same model as in the experiment of our main paper, namely a boosted forest with 10 trees. In Figure \ref{fig:boston_100} and Figure \ref{fig:boston_forest} the multilayer perceptron is trained for 100 epochs. In Figure \ref{fig:boston_200} it is trained for 200 epochs. We see that in all cases our approach outperforms the state-of-the-art and the baseline. Comparing Figure \ref{fig:boston_100} and \ref{fig:boston_200}, it is seems that our approach can also prevent overfitting.

\end{document}